\documentclass[runningheads]{llncs}
\usepackage{subfig}
\usepackage{url}
\usepackage{adjustbox}
\usepackage{float}
\usepackage{subfig}
\usepackage{cite}
\usepackage{multirow}
\usepackage{algorithm,algorithmic}
\usepackage{graphicx}
\usepackage{pgfplots}
\usepackage{tikz}
\usetikzlibrary{shapes,arrows}
\usetikzlibrary{positioning}
\usepackage{tkz-graph}
\usepackage{textcomp}
\usepackage{xcolor}
\usepackage{array}
\newcolumntype{C}[1]{>{\centering\let\newline\\\arraybackslash\hspace{0pt}}m{#1}}
\begin{document}
\title{\textit{What a Dialogue!}\\ A Deep Neural Framework for Contextual Affect Detection}
\titlerunning{A Deep Neural Framework for Contextual Affect Detection}
%
\author{Kumar Shikhar Deep \and Asif Ekbal \and Pushpak Bhattacharyya}
\authorrunning{K. Shikhar et al.}
%
\institute{Department of Computer Science and Engineering, \\ Indian Institute of Technology Patna, Patna, India \\
\email{\{shikhar.mtcs17,asif,pb\}@iitp.ac.in}}

%
\maketitle              
\begin{abstract}
A short and simple text carrying no emotion can represent some strong emotions when reading along with its context, i.e., the same sentence can express extreme anger as well as happiness depending on its context. In this paper, we propose a Contextual Affect Detection (CAD) framework which learns the inter-dependence of words in a sentence, and at the same time the inter-dependence of sentences in a dialogue. Our proposed CAD framework is based on a Gated Recurrent Unit (GRU), which is further assisted by contextual word embeddings and other diverse hand-crafted feature sets. Evaluation and analysis suggest that our model outperforms the state-of-the-art methods by 5.49\% and 9.14\% on Friends and EmotionPush dataset, respectively. 

\end{abstract}
\section{Introduction}
It becomes quite natural for us to gauge the emotion of a person if they explicitly mention that they are angry, sad or excited or even if they use the corresponding emojis at the end of the sentence, but what if there happens a drift in emotion while having a series of conversation between two people? And what if they stop using emotional emojis after a certain point of time even though they continue to be in the same state of emotion. Even human annotators may be confused if they do not consider context. 
Given that, even face-to-face conversation is confusing, sometimes, it should not be a matter of surprise if 
there could be a misinterpretation in textual conversations. The situation can get worse if there is a multi-party \footnote{Multi-Party conversation refers to the one having more than two speakers.} conversation. In this scenario, emotion of one speaker can change due to the utterance of the second speaker, and can again be switched due to the intervening of the third speaker. We have to be attentive to every speaker in the conversation or else our context would be lost.


Although a significant amount of research has been carried out for emotion analysis, only in the recent time there is a trend for performing emotion analysis of the dialogues in order to build an effective and human-like conversational agent. 

Our current work focuses on detecting emotions in a textual dialogue system. 
 We aim at labeling each utterance of a dialogue with one of the eight emotions, which comprises of Ekman's\cite{ekman1992there} six basic emotion tags, i.e., \textit{anger, fear, sadness, happiness, disgust} and \textit{surprise} plus \textit{neutral} and \textit{non-Neutral}\footnote{Non-Neutral emotion refers to the one having no majority voting of any one emotion type.} emotion. An example instance is depicted in Table \ref{fig:exm1}. If we look at the last utterance in the table i.e., \emph{There was no kangaroo!}, it can be considered as \textit{neutral} but while we consider its previous context, it should be assigned with the \textit{anger} class. 


\begin{table}[t]
\caption{Example utterances of a dialogue with their speaker and emotion label from EmotionLines\cite{chen2018emotionlines} 2018 dataset}
\centering
{
\resizebox{\textwidth}{!}
{
\begin{tabular}{l|l|l}
\hline
\textbf{Speaker} & \textbf{Utterance} & \textbf{Emotion} \\ \hline \hline 
Chandler & Good job Joe! Well done! Top notch! & Joy \\ \hline
Joey & You liked it? You really liked it? & Surprise \\ \hline
Chandler & Oh-ho-ho, yeah! & Joy \\ \hline
Joey & Which part exactly? & Neutral \\ \hline
Chandler & The whole thing! Can we go? & Neutral \\ \hline
Joey & Oh no-no-no, give me some specifics. & Non-Neutral \\ \hline
Chandler & I love the specifics, the specifics were the best part! & Non-Neutral \\ \hline
Joey & Hey, what about the scene with the kangaroo? Did-did you like that part? & Neutral \\ \hline
Chandler & I was surprised to see a kangaroo in a World War I epic. & Non-Neutral \\ \hline
Joey & You fell asleep!! & Anger \\ \hline
Joey & There was no kangaroo! & Anger \\ \hline
\end{tabular}
}
}
\label{fig:exm1}
\end{table}

Our deep neural network framework follows a stacking structure which utilizes bidirectional gated recurrent unit (Bi-GRU)
arranged in an hierarchical form. The lower Bi-GRU produces utterance level embeddings, and the upper-level Bi-GRU makes use of these embeddings to capture the contextual information in the dialogue. Some handcrafted features are incorporated at the different levels of the model to capture more linguistic evidences, 
which eventually found to be effective compared to the other models. We evaluate our proposed system on the benchmark dataset of EmotionLines 2018\cite{chen2018emotionlines}. We observe that our proposed framework attains better performance compared to the state-of-art model. 


\section{Related Work} 
\label{sec:lit}
Emotion detection and Sentiment classification have always been a hot research topic in the area of Natural Language Processing (NLP). 
Existing research on emotion detection have mostly focused on textual contents. 
In recent times, deep neural nets are being used very extensively to perform emotion analysis in a variety of domains, mediums and languages. 
Some of the most widely used models for capturing emotions include Convolutional Neural Network (CNN)\cite{lecun1995convolutional} and Recurrent Neural Networks (RNN)\cite{mikolov2010recurrent} like Long-Short Term Memory (LSTM)\cite{liu2016recurrent} and Gated Recurrent Unit (GRU)\cite{chung2014empirical}. 
All these works focus on to classifying emotions at the sentence level or utterance level, and thus cannot capture the context and inter-dependence among the utterances in dialogue.

\cite{poria2017context} proposed a network which made use of bidirectional long contextual short-term memory to detect the emotion of an utterance in dialogue and named it as bcLSTM. 
Later on, \cite{hazarika2018conversational} improved bcLSTM by introducing a memory network which makes use of speaker information as well for context modeling. 
The authors in \cite{luo2018emotionx} used Bi-LSTM to capture word dependencies and to extract relevant features for detecting various emotions. On top of that, they applied self-attention\cite{vaswani2017attention} to capture the inter-dependencies between the utterances of a dialogue. The work reported in \cite{saxena2018emotionx} uses hierarchical attention network model\cite{yang2016hierarchical} to embed contextual information among the utterances in a dialogue. 
\cite{jiao2019higru} used a bidirectional gated recurrent unit (Bi-GRU)\cite{chung2015gated} fused with self-attention\cite{vaswani2017attention} and its word embeddings to efficiently utilize word-level as well as utterance-level information. 

Our proposed model differs from the existing models in the sense that we derive deep contextualized representation of each word in an utterance, and then incorporate it into the model as a feature along with the pre-trained Glove word embedding\cite{pennington2014glove}. We acquire these word embedding from the pre-trained ELMo\cite{Peters:2018} model. These representations take the entire context into account. Being character based, they allow the network to use morphological cues to form robust representations for out-of-vocabulary words unseen in training. We use hierarchical Bi-GRU to learn the context in a dialogue fused with various handcrafted features obtained through transfer learning over similar tasks.

 \begin{figure*}[!ht]
     \centering
     \includegraphics[width=\textwidth]{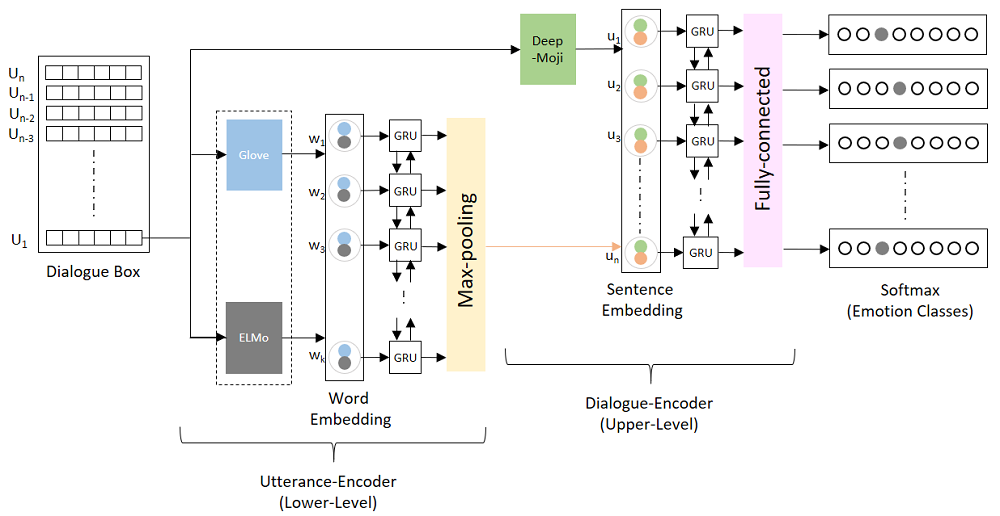}
     \caption{Proposed Architecture}
     \label{fig:arch}
 \end{figure*}
 
\section{Proposed Methodology} \label{sec:method}
In this section, we describe our proposed framework in details. Unlike the previous models, which performed exceptionally well in predicting some emotions follows complex architecture.. 
Evaluation shows that a model which performs very well for 
a specific emotion class 
compensates with the lower performance in other 
emotion classes (i.e. performs at the cost of other classes). In contrast, our model is straightforward and efficient, and at the same time outperforms every different models with significant margins. It consists of two layers of hierarchy- the lower layer is for encoding the utterances (named as \textit{utterance-encoder}), and the upper layer is for encoding the dialogues (named as \textit{dialogue-encoder}).

Given a batch/dialogue\footnote{We treat each dialogue as one batch}, firstly the individual utterances are passed through utterance-encoder which comprises of a recurrent layer (biGRU) followed by a max-pooling layer. GRU learns the contextual representation of each word in an utterance, i.e., the representation of each word is learned based on the sequence of words in the utterance. Subsequently, we apply max-pooling over the hidden representation of each utterance to capture the most important features over time. The obtained utterance representations form the input to the dialogue-encoder, which again comprises of biGRU to capture the contextual information of each utterance in a dialogue. Since the task is to classify each utterance of a dialogue, the hidden representations of biGRU over the time are passed through fully connected layer followed by Softmax to obtain the corresponding emotion label. Further, the inputs to utterance encoder and dialogue encoder are assisted by a diverse set of hand-crafted features (c.f. Section \ref{sec:feat}) for the final prediction. Figure \ref{fig:arch} depicts our proposed architecture.

\subsection{Hand-crafted Features}
\label{sec:feat}
We make use of transfer learning to capture important evidences obtained from the various state-of-the-art pre-trained deep learning models. 
Following sub-sections explain these models in details:
\begin{itemize}
    \item \textbf{DeepMoji} \cite{felbo2017}: DeepMoji performs distant supervision on a very large dataset \cite{thelwall2010sentiment} 
    (1.2 billion tweets) comprising of noisy labels (emojis). By incorporating transfer learning on various downstream tasks, they were able to outperform the state-of-the-art results on 8 benchmark datasets on 3 NLP tasks across the five domains. DeepMoji can give an excellent representation of text which can be incorporated in any sentiment or emotion detection model to improve the performance. Since our target task is closely related to this, we adopt this for our domain and extract the embeddings of 2304 dimension from the attention layer, which acts as the utterance embedding feature for upper-level (dialogue-encoder) of the model.  

    \item \textbf{ELMo} \cite{Peters:2018}: Unlike traditional word embedding techniques, ELMo makes use of bidirectional LSTM network to create word representations. This biLSTM is trained with a coupled language model objective on a large text corpus. Thus we can say, each word representation is a function of the entire sentence. It analyses the words within the context that they are used, thus capturing the syntactic as well as semantic characteristics of the words and also take care of variance across the linguistic contexts. Also, being character based, we can have representations of out-of-vocabulary words as well. Since it is proven that the addition of ELMo representations can improve the performance of any NLP model, we incorporate it into our model along with the pre-trained Glove word embedding. To reduce the processing time, we extracted the 1024 dimensional ELMo word embeddings beforehand instead of creating it during the process. 
 \end{itemize}

\subsection{Word Embedding} \label{sec:we}
Embedding matrix is generated from the pre-processed text using a combination of pre-trained and deep contextualized word embedding:
\begin{enumerate}
    \item \textbf{Pre-trained GloVe embeddings for tweets} \cite{pennington2014glove}: We use 300-dimensional pre-trained GloVe word embeddings, trained on the Twitter corpus, for the experiments. The glove is a count-based model that captures the count of the word, i.e., how frequently it appears in a context.
    \item \textbf{Deep contextualized ELMo embeddings} \cite{Peters:2018}: We extract 2304 dimensional word vectors from ELMo embedding model, which learns the embedding from an internal state of biLSTM and represents word-level contextual features of the input text. 
    
\end{enumerate}
We finally concatenate the word representation of both the embeddings, which act as input to the utterance-encoder (lower level).

\section{Experiments and Results}
\label{sec:exp}

\begin{table}[]
\caption{Statistics of EmotionLines 2018\cite{chen2018emotionlines} dataset }
\label{fig:exm2}
\centering
{
\resizebox{\linewidth}{!}
{
\begin{tabular}{|c|c|c|c|c|c|c|c|c|c|c|c|}
\hline
\multirow{2}{*}{\textbf{Dataset}} & \multicolumn{3}{c|}{\textbf{\begin{tabular}[c]{@{}c@{}}\#Dialogues\\ (\#Utterances)\end{tabular}}} & \multicolumn{8}{c|}{\textbf{Emotion Label Distribution}} \\ \cline{2-12} & \textit{Train} & \textit{Validation} & \textit{Test} & \textit{Neu} & \textit{Joy} & \textit{Sad} & \textit{Fea} & \textit{Ang} & \textit{Sur} & \textit{Dis} & \textit{Non} \\ \hline 
\textbf{Friends} & \begin{tabular}[c]{@{}c@{}}720\\ (10,561)\end{tabular} & \begin{tabular}[c]{@{}c@{}}80\\ (1,178)\end{tabular} & \begin{tabular}[c]{@{}c@{}}200\\ (2,764)\end{tabular} & 6530 & 1710 & 498 & 246 & 759 & 1657 & 331 & 2772 \\ \hline
\textbf{EmotionPush} & \begin{tabular}[c]{@{}c@{}}720\\ (10,733)\end{tabular} & \begin{tabular}[c]{@{}c@{}}80\\ (1,202)\end{tabular} & \begin{tabular}[c]{@{}c@{}}200\\ (2,807)\end{tabular} & 9855 & 2100 & 514 & 42 & 140 & 567 & 106 & 1418 \\ \hline
\end{tabular}
}
}
\end{table}

\subsection{Dataset Description}
For experiments, 
we use the benchmark dataset of EmotionLines 2018\cite{chen2018emotionlines}, which is an emotion annotated corpus of multi-party conversation. The dataset comprises of two individual corpus of dialogue set extracted from two different sources, one from the famous TV show scripts named Friends, and the other from human to human chat logs on Facebook messenger through an application called EmotionPush. 

\begin{enumerate}
\item \textbf{Friends TV series data :}
The Friends script was crawled, and each scene of an episode was treated as a dialogue. Thus each dialogue consists of multiple speakers. \cite{chen2018emotionlines} separated the dialogues based on its window size\footnote{Window size refers to number of utterances in a dialogue} of [5, 9], [10, 14], [15, 19], and [20, 24]. Finally, they randomly collected 250 dialogues from each- thus creating a corpus of 1000 dialogues, which are further splitted up into 720, 80, and 200 dialogues for training, validation and testing, respectively.
\item \textbf{EmotionPush Chat log data:}
This data was collected by crawling the private conversation among the friends on facebook messenger with the help of EmotionPush app. To protect the private information of the users like names, organizations, locations, email address, and phone numbers, they used a two-step masking procedure. They treated the conversations lasting not more than 30 minutes as a dialogue. Finally, they make use of the same procedure for sampling and categorizing as they used for Friends TV script and collected 1000 dialogues which were again divided in the same ratio for training, validation, and testing. 
\end{enumerate}

\begin{figure}[ht!]
	\centering
	    \subfloat[Friends dataset \label{fig:data:class1}]
 	    {%
            \resizebox{0.46\textwidth}{!}
            {
 			    \begin{tikzpicture}
                    \begin{axis}[
                        name=plot1,
                        ybar,
                        enlargelimits=0.15,
                        width=\textwidth,
                        height=8.5cm,
                        bar width=0.2cm,
                        ylabel ={Number of labels},
                        ylabel near ticks,
                        xlabel={Emotion Classes},
                        symbolic x coords={Neu, Joy, Sad, Fea, Ang, Sur, Dis, Non},
                        xtick=data,
                        area legend,
                        legend style={font=\scriptsize, legend pos= north east, legend columns=1,  legend style={row sep=0.5pt}},
                        ]
                        \addplot[color=red, fill=red!50,] coordinates {(Neu,4752) (Joy,1283) (Sad,351) (Fea,185) (Ang,513) (Sur,1220) (Dis,240) (Non,2017)};\addlegendentry{Train - 10,561}
                        \addplot[color=cyan, fill=cyan!50,] coordinates {(Neu,491) (Joy,123) (Sad,62) (Fea,29) (Ang,85) (Sur,151) (Dis,23) (Non,214)};\addlegendentry{Validation - 1178}
                        \addplot[color=blue, fill=blue!50,] coordinates {(Neu,1287) (Joy,304) (Sad,85) (Fea,32) (Ang,161) (Sur,286) (Dis,68) (Non,541)};\addlegendentry{Test - 2764}
                    \end{axis}
                \end{tikzpicture}
     	    }
        }
        \hspace{2em}
        \subfloat[EmotionPush dataset \label{fig:data:class2}]
 	    {%
            \resizebox{0.46\textwidth}{!}
            {
 			    \begin{tikzpicture}
                    \begin{axis}[
                        name=plot1,
                        ybar,
                        enlargelimits=0.15,
                        width=\textwidth,
                        height=8.5cm,
                        bar width=0.2cm,
                        ylabel ={Number of labels},
                        ylabel near ticks,
                        xlabel={Emotion Classes},
                        symbolic x coords={Neu, Joy, Sad, Fea, Ang, Sur, Dis, Non},
                        xtick=data,
                        area legend,
                        legend style={font=\scriptsize, legend pos= north east, legend columns=1,  legend style={row sep=0.5pt}},
                        ]
                        \addplot[color=red, fill=red!50,] coordinates {(Neu,1064) (Joy,1482) (Sad,389) (Fea,36) (Ang,94) (Sur,435) (Dis,85) (Non,1064)};\addlegendentry{Train - 10,733}
                        \addplot[color=cyan, fill=cyan!50,] coordinates {(Neu,825) (Joy,160) (Sad,38) (Fea,4) (Ang,9) (Sur,39) (Dis,6) (Non,121)};\addlegendentry{Validation - 1202}
                        \addplot[color=blue, fill=blue!50,] coordinates {(Neu,1882) (Joy,458) (Sad,87) (Fea,2) (Ang,37) (Sur,93) (Dis,15) (Non,233)};\addlegendentry{Test - 2807}
                    \end{axis}
                \end{tikzpicture}
     	    }
        }
        \caption{Emotion class distribution of Friends and EmotionPush dataset}
        \label{fig:data}
\end{figure}
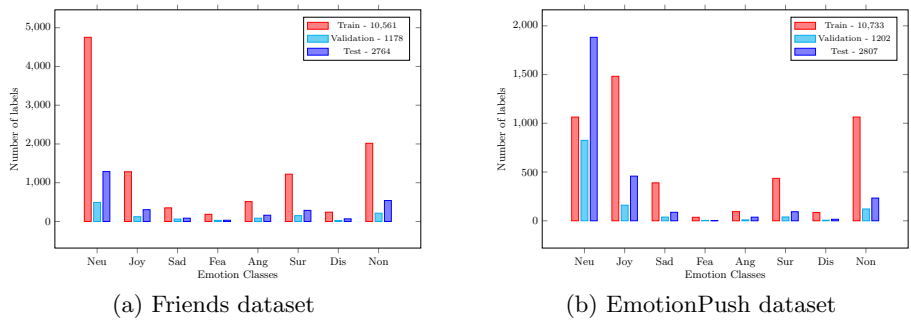

\noindent Table \ref{fig:exm2} shows the distribution of both Friends and EmotionPush datasets in terms of the number of dialogues, the number of utterances, and the number of emotion labels. To compare our model we follow the setup of \cite{hsu2018socialnlp} and evaluate its performance only on four emotions, i.e., anger, joy, sadness and neutral on both Friends and EmotoinPush datasets, and excluding all the other emotion classes during training of our model. We ignore other emotion classes by setting their corresponding loss weights to zero. Fig \ref{fig:data} depicts the distribution of emotion classes into train, validation, and test set for both Friends and EmotionPush datasets.

\tikzstyle{in-out}=[draw, fill=red!20, text width=15em, outer sep=2, text centered, minimum height=3em, rounded corners]
\tikzstyle{in-out-large}=[draw, fill=green!30, text width=8em, outer sep=2, text centered, minimum height=5em, rounded corners]
\tikzstyle{process}=[draw, outer sep=2, fill=teal!20, text width=31.5em, text centered, minimum height=3em]
\tikzstyle{process-2}=[draw, outer sep=2, fill=green!30, text width=31.5em, text centered, minimum height=3em, rounded corners]
\tikzstyle{process-small}=[draw, outer sep=2, fill=blue!20, text width=15em, text centered, minimum height=2.5em]
\tikzstyle{process-large}=[draw, outer sep=2, fill=blue!20, text width=31.5em, text centered, minimum height=4em]
\tikzstyle{process-huge}=[draw, outer sep=2, fill=blue!20, text width=38em, text centered, minimum height=6em]

\tikzstyle{in-out-new}=[draw, fill=red!20, text width=5em, outer sep=2, text centered, minimum height=3em, rounded corners]
\tikzstyle{process-2-new}=[draw, outer sep=2, fill=green!30, text width=12em, text centered, minimum height=3em, rounded corners]
\tikzstyle{process-small-new}=[draw, outer sep=2, fill=blue!20, text width=5em, text centered, minimum height=2.5em]
\tikzstyle{process-large-new}=[draw, outer sep=2, fill=blue!20, text width=12em, text centered, minimum height=4em]
\tikzstyle{process-new}=[draw, outer sep=2, fill=teal!20, text width=41.0em, text centered, minimum height=3em]

\tikzstyle{in-out-dummy}=[text width=3em, outer sep=0, text centered, minimum height=3em, rounded corners]

\subsection{Pre-processing}
Friends data consists of scene snippets containing multi-speaker conversation, while EmotionPush data includes Facebook messenger chats between two individuals. Both datasets contain some incomplete sentences and excessive use of punctuations. In addition to it, EmotionPush data also contains emoticons which are absent in the Friends data. We believe that the reason behind it is that Friends data is the script which is collected by converting audio to text. We perform the following pre-processing steps: 

(i). all the characters in text are converted to lower case; (ii). remove punctuation symbols except ! and ? because we believe '!' and '?' may contribute to a better understanding of intense emotions like surprise and anger; and (iii). remove extra space, emoticons and the newline character. 
Finally, we perform word tokenization to construct the vocabulary of words and symbols. The tokens, thus, collected are mapped to their corresponding 300-dimensional Glove vectors.

\subsection{Experiments}
We pad each utterance to a maximum length of 50 words. The GRU dimension is set to 300 for both lower and upper level. We employ 300 plus 1024 dimensional word embeddings and 300 plus 2304 dimensional sentence embeddings for the experiments. We use \textit{Tanh} activation and set the \textit{Dropout} \cite{dropout} as 0.5 in order to prevent the model from overfitting. We optimize our model using \textit{Adam} 
optimizer along with \textit{weighted categorical cross-entropy} loss functions for emotion classification. From Table \ref{fig:exm2}, it is clear that EmotionLines data suffers from class imbalance issue. We follow \cite{khosla2018emotionx} to prevent our model from getting biased towards more frequently occurring emotion classes-thereby providing larger weights to the losses corresponding to less frequently occurring classes and vice-versa.




We conduct our experiments on the Pytorch framework\footnote{\textit{Pytorch Home Page.} \url{http://www.pytorch.org/.}}. We adopt the official evaluation metric of EmotionX 2018\cite{hsu2018socialnlp} shared task, i.e. weighted accuracy (WA) (eq.\ref{wa}) and un-weighted accuracy (UWA) (eq.\ref{uwa}), for measuring the performance of our model. We train our model for the maximum 50 epochs with early stopping criteria on validation accuracy, having the patience of 10. We initialize the learning rate by 0.00025 with the decaying factor of 0.5 on every 15 epochs.
\begin{equation}
\label{wa}
        WA = \sum_{c \epsilon C}{p_c . a_c}
\end{equation}
\begin{equation}
\label{uwa}
        UWA = \frac{1}{|C|}\sum_{c \epsilon C}{a_c}
\end{equation}
where $a_c$ refers to the accuracy of emotion class c and $p_c$ refers to the percentage of emotion class c in the test set.

\begin{table}[]
\caption{Experimental results on Friends and EmotionPush datasets. F(E) denotes that the training is done on the corresponding dataset while F+E indicates that training is done on both Friends and EmotionPush dataset.\\}
\centering
\resizebox{\textwidth}{!}{%
\begin{tabular}{|c|c|c|c|c|c|}
\hline
& & \multicolumn{2}{c|}{\textbf{Friends Dataset}} & \multicolumn{2}{c|}{\textbf{EmotionPush Dataset}} \\ \cline{3-6} 
\multirow{-2}{*}{\textbf{\begin{tabular}[c]{@{}c@{}}Framework\\ Used\end{tabular}}} & \multirow{-2}{*}{\textbf{\begin{tabular}[c]{@{}c@{}}Train\\ Data\end{tabular}}} & 
\textit{WA} & \textit{UWA} & \textit{WA} & \textit{UWA} \\ 
\hline
\textbf{HiGRU}\cite{jiao2019higru} & F(E) & \textbf{74.4} & 67.2 & 73.8 & 66.3 \\ 
\hline
\textbf{HiGRU-sf}\cite{jiao2019higru} & F(E) & 74 & \textbf{68.9} & 73 & 68.1 \\ 
\hline
\textbf{HiGRU-sf} & F+E & 69 & 64.8 & \textbf{77.1} & \textbf{70.2} \\ 
\hline
\textbf{CAD} & F(E) & {\color[HTML]{9A0000} 75.94} & {\color[HTML]{9A0000} 74.39} & {\color[HTML]{9A0000} 86.24} & {\color[HTML]{9A0000} 80.18} \\ 
\hline
\multicolumn{2}{|l|}{\textbf{Class Accuracies}} & \multicolumn{2}{l|}{\begin{tabular}[c]{@{}l@{}}\textit{Neu}: 75.14  \textit{Joy}: 83.88\\ \textit{Sad}: 65.88  \textit{Ang}: 72.67\end{tabular}} & 
\multicolumn{2}{l|}{\begin{tabular}[c]{@{}l@{}}\textit{Neu}: 87.62  \textit{Joy}: 83.84\\ \textit{Sad}: 73.56  \textit{Ang}: 75.68\end{tabular}} \\ 
\hline
\end{tabular}%
}
\label{tab:results}
\end{table}


Table \ref{tab:results} shows the evaluation results of the top-performing existing models so far compared with that of our proposed framework. In the table, HiGRU and HiGRU-sf are proposed by \cite{jiao2019higru}. Results show that our model outperforms all the other models with a significant margin. Further, we observe the improvement to be statistically significant with 95\% and 99\% confidence on Friends and EmotionPush dataset, i.e., the p-value is less than 0.05 for paired T-test\cite{kim2015t} of both the datasets. On Friends dataset, our model reports un-weighted and weighted accuracy of 74.39\% and 75.94\% as compared to that of 68.9\% and 74.4\% of the state-of-art models, thus improving them by 5.49\% and 1.54\%. On the other hand, our CAD framework performs better on EmotionPush dataset as well, reporting the unweighted and weighted accuracies of 80.18\% and 86.24\%, respectively, as compared to that of 70.2\% and 77.1\% in the state-of-art model, thus giving an increment of 9.14\% and 9.98\%. It is worth noticing that the  accuracy of anger in EmotionPush dataset increases by 17.38 points without compensating with the other classes. This implies that for all the emotion classes our framework performs reasonably competent with the accuracies being balanced and not biased to any specific class.

\begin{table}[]
\caption{A set of two dialogues from Friends' test set showing the correct predictions of rare emotions}
\centering
\resizebox{\textwidth}{!}{%
\begin{tabular}{@{}|c|l|c|c|@{}}
\hline
\textbf{Speaker} & \multicolumn{1}{c|}{\textbf{Utterance}} & \textbf{True Emotion} & \textbf{Predicted Emotion} \\ \hline \hline
Nurse & This room's available. & Neutral & Neutral \\ \hline
Rachel & \begin{tabular}[c]{@{}l@{}}Okay!\end{tabular} & Joy & Joy \\ \hline
 & \begin{tabular}[c]{@{}l@{}}: \\ : \\\end{tabular} &  &  \\ \hline
Rachel & You listen to me! & Anger & Anger \\ \hline \hline
Chandler & Hey! & Joy & Neutral \\ \hline
Monica & \begin{tabular}[c]{@{}l@{}}Hi!\end{tabular} & Neutral & Neutral \\ \hline
 & \begin{tabular}[c]{@{}l@{}}: \\ : \\\end{tabular} &  &  \\ \hline
Chandler & So, I guess this is over. & Sadness & Sadness \\ \hline
\end{tabular}%
}
\label{tab:examples}
\end{table}
To test the correctness of our prediction, we adopt some examples where the emotion drift occurs over an extended context. It can be seen from the Table \ref{tab:examples} that rare emotions are predicted correctly. Thus we can say that our model captures the contextual information in the dialogue pretty well.

\subsection{Error Analysis}
\begin{figure}[ht!]
    \centering
    \subfloat[Friends]{\includegraphics[width=0.46\textwidth]{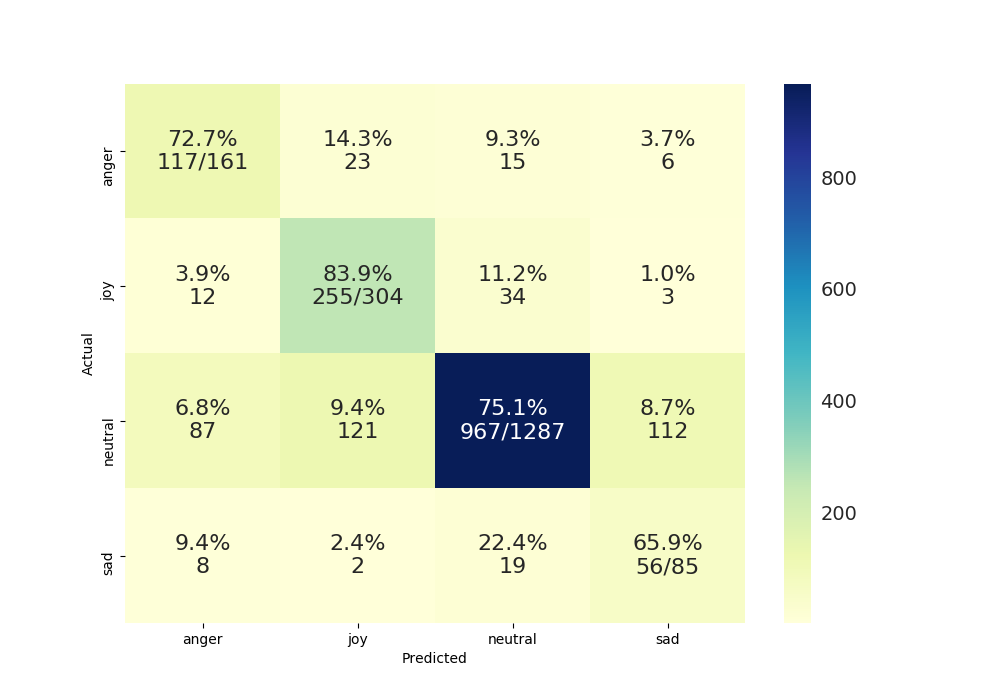}\label{fig:confusion:mtl:dl}}
    \hspace{0.5cm}
     \subfloat[EmotionPush]
     {\includegraphics[width=0.46\textwidth]{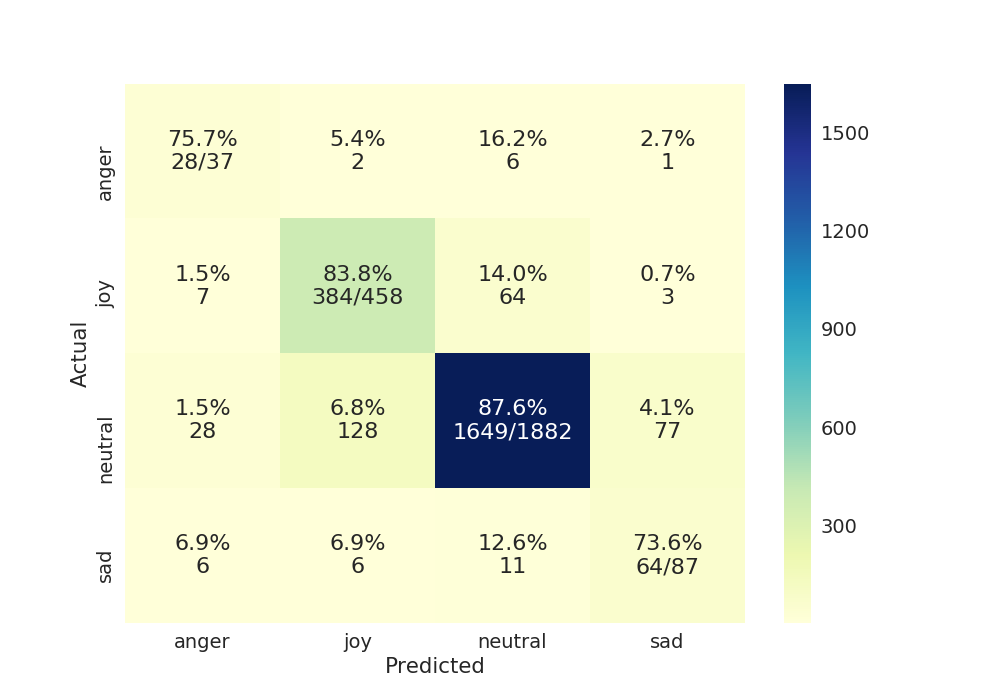}\label{fig:confusion:mtl:ml}}
     \caption{Confusion matrices for Emotion Classification.}
     \label{fig:confusion}
\end{figure}

 In Figure \ref{fig:confusion}, we show the confusion matrices for both the datasets (i.e., Friends and EmotionPush) for quantitative analysis. 
We find that \textit{joy} and \textit{sad} emotions are mostly confused with \textit{neutral}, possibly due to the absence of affective words and large number of \textit{neutral} class.
Most of the \textit{neutral} emotions get confused with \textit{joy} due to the presence of some positive sentiment words.
Majority of \textit{anger} emotion in EmotionPush dataset are missclassified as \textit{neutral} (mostly due to absence of any sentiment bearing words) while that of Friends dataset are misclassified as \textit{joy} due to presence of exclaimation mark which shows strong emotion such as \textit{joy} or \textit{anger}. The presence of positive sentiment enforce it to predict \textit{joy}.

\noindent We also perform a qualitative analysis to get better insights about the strength and weakness of the system. We found that, apart from dominating the effect of higher distribution class, few other error cases are present such as:
\begin{itemize}
    \item \textbf{Loss of emotion drift:} When a strong emotion such as \textit{anger} is encountered, it continues to be predicted. The reason behind it might be because we did not consider the speaker information while extracting contextual information from the utterances. An example is shown in the first part of Table \ref{tab:er}. 
    \item \textbf{Usage of expression like \textit{lol, haha, etc.}:} Conversations involving such expressions have led to misclassifications of prediction by the model for some of the instances. An example is given in the second part of Table \ref{tab:er}. The sentences are originally labeled \textit{Sad}. However, due to the presence of a smiling expression-related term (\textit{haha} and \textit{lol}), the model has predicted it as \textit{Joy}.

\end{itemize}

\begin{table}[]
\caption{Table showing loss of emotion drift in a dialogue of Friends dataset.}
\centering
\resizebox{\textwidth}{!}{%
\begin{tabular}{@{}lcc@{}}
\hline
\multicolumn{1}{c}{\textbf{Utterances}} & \textbf{True} & \textbf{Predicted} \\ \hline
Don't ask me, I had it and I blew it! & Anger & Anger \\
Well, I want it! & Other & Anger \\
You can have it! & Other & Anger \\
I don't know, maybe I can't. I mean, maybe there's something wrong with me. & Other & Anger \\
Oh, no! No! & Other & Anger \\
It's out there man! I've seen it! I got it!! & Joy & Anger \\
Then you hold on to it!! & Other & Anger \\
All right, man!! & Joy & Anger \\
All right, congratulations you lucky bastard! & Joy & Joy \\ \hline \hline
I got it wrong lol & Sad & Joy \\
gone but not forgotten haha & Sad & Joy \\
\hline
\end{tabular}%
}
\label{tab:er}
\end{table}

\rmfamily

\section{Conclusion}\label{sec:con}
In this paper, we have presented a hybrid deep neural network framework (CAD) for detecting emotions in a dialogue. We propose a hierarchical BiGRU network which takes the assistance of various hand-crafted feature set on different levels of architecture to learn the contextual information of dialogue. The learned representation is fed to a fully connected layer over the time steps followed by softmax layer for the predictions. We have evaluated our model on the benchmark datasets of EmotionLines-2018, which consists of two corpora i.e., Friends and EmotionPush data. The evaluation suggests that our CAD framework obtains improved results against the state-of-art model so far. This model can be applied to various other similar datasets as well to improve their results.

\section{Acknowledgement}
Asif Ekbal acknowledges the Young Faculty Research Fellowship (YFRF), supported by Visvesvaraya Ph.D. scheme for Electronics and IT, Ministry of Electronics and Information Technology (MeitY), Government of India, being implemented by Digital India Corporation (formerly Media Lab Asia).

\bibliographystyle{splncs04}
\bibliography{mybibliography}

\end{document}